\documentclass[letterpaper]{article}

\usepackage{spconf,amsmath,graphicx}
\usepackage{amssymb}
\usepackage{multirow}
\usepackage{url}
\usepackage{hyperref}
\hypersetup{hidelinks}
\usepackage[table,xcdraw]{xcolor}
\usepackage{subcaption}
\usepackage{adjustbox}

\title{An Object-Centered Data Acquisition Method for 3D Gaussian Splatting using Mobile Phones}
\name{%
Yuezhe Zhang$^{1}$ \qquad
Luqian Bai$^{1}$ \qquad
Mengting Yu$^{1}$ \qquad
Lei Wei$^{2}$ \qquad
Shuai Wan$^{1}$ \qquad
Yifan Zhang$^{1}$
}

\address{%
$^{1}$School of Electronics and Information, Northwestern Polytechnical University, Xi'an, China\\
$^{2}$College of Engineering, Xi'an International University, Xi'an, China
}

\begin{document}
%
\maketitle
\begin{abstract}
 Data acquisition through mobile phones remains a challenge for 3D Gaussian Splatting (3DGS). In this work we target the object-centered scenario and enable reliable mobile acquisition by providing on-device capture guidance and recording onboard sensor signals for offline reconstruction. After the calibration step, the device orientations are aligned to a baseline frame to obtain relative poses, and the optical axis of the camera is mapped to an object-centered spherical grid for uniform viewpoint indexing. To curb polar sampling bias, we compute area-weighted spherical coverage in real-time and guide the user's motion accordingly. We compare the proposed method with RealityScan and the free-capture strategy. Our method achieves superior reconstruction quality using fewer input images compared to free capture and RealityScan. Further analysis shows that the proposed method is able to obtain more comprehensive and uniform viewpoint coverage during object-centered acquisition. An anonymized implementation is available at:
\url{https://zyz-nwpu.github.io/3dgs-oc-capture/}

\end{abstract}
\section{Introduction}
\label{sec:intro}

3D Gaussian Splatting (3DGS) \cite{b1} has recently emerged as a promising neural scene representation that strikes an effective balance between high rendering fidelity and real-time performance \cite{b2}. By modeling scenes with anisotropic Gaussian primitives and leveraging differentiable rasterization, 3DGS enables a real-time training-and-rendering pipeline \cite{b3}, thereby mitigating the training and rendering issues of prior approaches such as NeRF \cite{b4}. However, the performance of 3DGS is highly sensitive to the quality and distribution of input data, where insufficient scene coverage and lack of camera pose information generally lead to artifacts \cite{b5}, missing regions, and structural inconsistencies \cite{b6}.

\begin{figure}[!t]
  \centering
  \includegraphics[width=1\columnwidth]{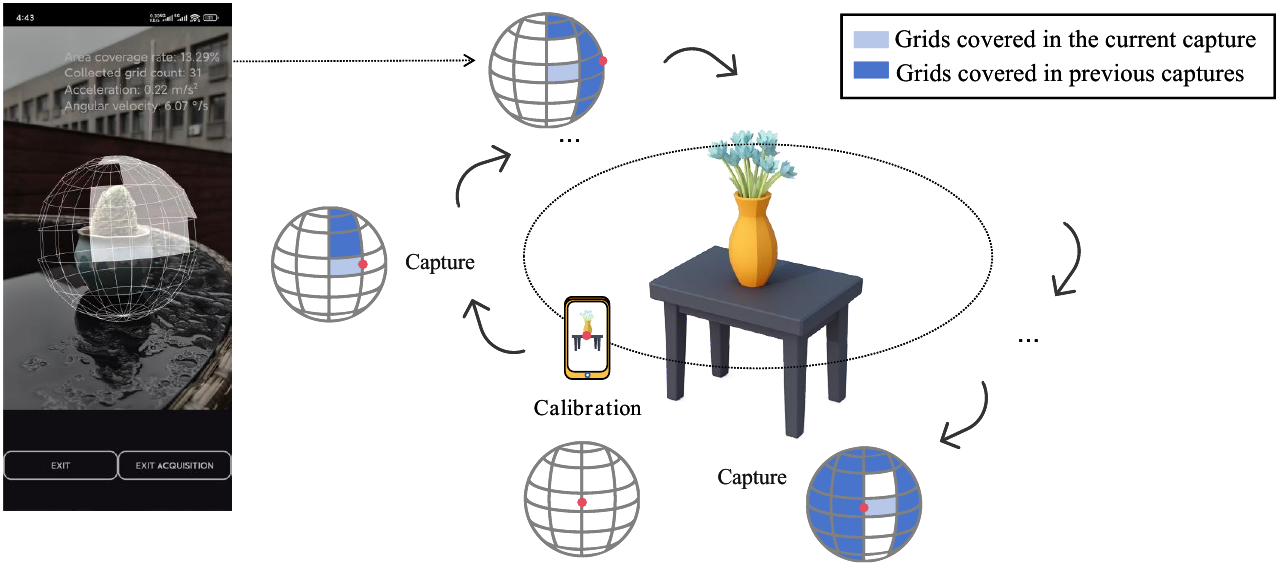} 
  \caption{Mobile acquisition pipeline for 3DGS with object-centered calibration and grid-guided capture for spherical viewpoint coverage.
}
  \label{fig:example1}
\end{figure}

Efficiently acquiring dense, high-quality multi-view data remains a central challenge for 3DGS applications. Typical 3DGS acquisition using static camera arrays or mechanical rigs can provide precise control, but those methods are costly, inflexible, and difficult to deploy in dynamic environments. In contrast, mobile  phones equipped with integrated Inertial Measurement Unit (IMU) sensors and high-resolution cameras offer an easy-to-deploy, low-cost alternative~\cite{b7}~\cite{b8}. However, mobile capture introduces challenges including pose instability, uneven viewpoint coverage, and motion-induced blur, which can degrade reconstruction quality if not well addressed~\cite{b9}.

To tackle these issues, we propose a spherical-viewpoint acquisition method tailored to 3DGS acquisition using mobile phones, as illustrated in Fig.~\ref{fig:example1}. The proposed method establishes an object-centered spherical coordinate framework and uses real-time IMU-based orientation estimation to map camera views onto spherical coordinates~\cite{b10}. By continuously evaluating coverage on a discretized spherical grid and monitoring device motion, the proposed method guides the user toward complete and uniform multi-view acquisition under steady conditions, improving the quality.
\newlength{\figxshift}      
\setlength{\figxshift}{15pt}

\begin{figure*}[!t]
  \centering
  \makebox[\linewidth][l]{\hspace*{\figxshift}%
    \includegraphics[width=\linewidth]{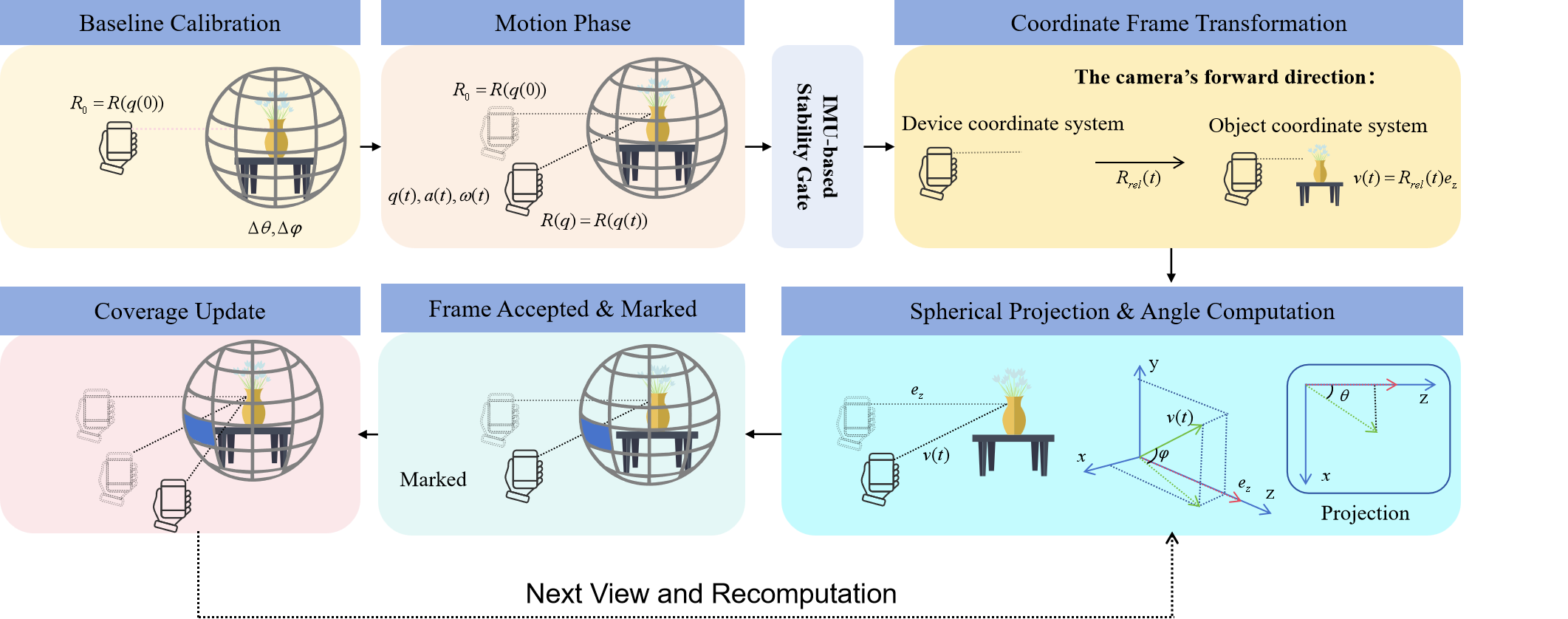}}
  \caption{
  Object-centered spherical coordinate mapping and coverage update workflow. The baseline rotation $R_0 = R(q(0))$ is obtained from the initial quaternion $q(0)$. During data acquisition, the device records quaternion $q(t)$, acceleration $a(t)$, and angular velocity $\omega(t)$; after IMU-based stability gate, the rotation matrix $R(q(t))$ and relative rotation $R_{\mathrm{rel}}(t) = R_0^{\top}R(t)$ are derived. The camera forward vector $e_z = [0,0,1]^{\top}$ is transformed to the object frame as $v(t) = R_{\mathrm{rel}}(t)e_z$ and converted into spherical coordinates $\theta$ and $\phi$. Each stable capture is mapped to a sampling direction $(p, t)$ on the spherical grid, with the corresponding cell $C(p, t)$ used to update the area-weighted coverage in real time.
  }
  \label{fig:111}
\end{figure*}

The main contributions of this work are as follows:

1) We map orientations estimated from mobile phones' IMU to an object-centered spherical coordinate system.

2) We provide real-time feedback of area-weighted spherical coverage to guide users during data acquisition to improve angular uniformity and completeness.

3) We introduce a dual-mode stability gate based on smoothed linear acceleration and angular velocity, to ensure only frames under steady-motion conditions are used.

Experimental results demonstrated that the proposed method enables high-quality reconstruction of 3DGS, in terms of image quality and viewpoint coverage. The following section details the pose method and spherical computation.

\section{Camera Pose Modeling and Spherical Coordinate Computation}
\label{sec:format}

As shown in Fig.~2, the pipeline performs IMU-based pose acquisition, spherical mapping, and online coverage. Inputs are \( \mathbf{q}(t), \mathbf{a}(t), \boldsymbol{\omega}(t) \). After stability gate, \( \mathbf{q}(t) \!\rightarrow\! R(\mathbf{q}) \), aligned with \( R_0 \) to yield \( R_{\mathrm{rel}} \)~\cite{b11}; the viewing direction is \( \mathbf{v}=R_{\mathrm{rel}} e_z \) with angles \( (\theta,\phi) \). Each accepted frame updates its spherical grid cell \( C(p,t) \), and area-weighted coverage is accumulated to enforce uniform viewpoints. All parameters are empirically determined based on preliminary experiments.

The following subsections describe the two main components of this process.

\subsection{Angle Calculation of Viewing Direction using Relative Rotation Matrix}

During operation, the camera continuously captures images while recording motion data from its built-in inertial measurement unit.
The rotation-vector sensor provides the device orientation as a quaternion $\mathbf{q}(t) = [q_x, q_y, q_z, q_w]^{\top}$, while the accelerometer and gyroscope measure linear acceleration~ $\mathbf{a}(t) = (a_x, a_y, a_z)$ and angular velocity $\boldsymbol{\omega}(t) = (\omega_x, \omega_y, \omega_z)$, respectively.

To reduce short-term sensor noise and jitter, the magnitudes $\|\mathbf{a}\|(t)$ and $\|\boldsymbol{\omega}\|(t)$ are smoothed using an exponential moving average (EMA)~\cite{b12}:
\begin{align}
\widehat{a}(t_k) &= \alpha \, \widehat{a}(t_{k-1}) + (1-\alpha)\, \|\mathbf{a}\|(t_k), \\
\widehat{\omega}(t_k) &= \alpha \, \widehat{\omega}(t_{k-1}) + (1-\alpha)\, \|\boldsymbol{\omega}\|(t_k).
\end{align}
A pose is considered stable only when
\begin{align}
\widehat{a}(t_k) \leq a_{\mathrm{th}}, \quad \widehat{\omega}(t_k) \leq \omega_{\mathrm{th}},
\end{align}
and this condition holds over a time window $\Delta t_{\mathrm{hold}}$.
Otherwise, the interval is marked unstable, and both computation and data acquisition are suspended.

Once stability is verified, the quaternion $\mathbf{q}(t)$ is converted into a direction cosine matrix (DCM) $R(t) \in \mathrm{SO}(3)$~\cite{b13}, which maps vectors from the device frame to the world frame while preserving orthogonality and unit determinant:
\begin{align}
R(q) = \scalebox{0.85}{$
\begin{bmatrix}
1-2(q_{y}^{2}+q_{z}^{2}) & 2(q_{x} q_{y}-q_{z} q_{w}) & 2(q_{x} q_{z}+q_{y} q_{w}) \\
2(q_{x} q_{y}+q_{z} q_{w}) & 1-2(q_{x}^{2}+q_{z}^{2}) & 2(q_{y} q_{z}-q_{x} q_{w}) \\
2(q_{x} q_{z}-q_{y} q_{w}) & 2(q_{y} q_{z}+q_{x} q_{w}) & 1-2(q_{x}^{2}+q_{y}^{2})
\end{bmatrix}$}.
\end{align}

Because the initial camera placement may vary across experiments, directly using $R(t)$ can yield inconsistent references. To maintain comparability, the baseline orientation $R_0$ is recorded at the start of acquisition, and subsequent orientations are expressed as relative rotations:
\begin{align}
R_{\mathrm{rel}}(t) = R_0^\top R(t),
\end{align}

Then the viewing direction is \(\mathbf{v}(t)=R_{\mathrm{rel}}(t)\mathbf{e}_z\) with \(\mathbf{e}_z=[0,0,1]^{\top}\). By the standard Cartesian–to–spherical mapping in \(\mathbb{R}^3\), for \(\mathbf{v}(t)=(v_x,v_y,v_z)^{\top}\) , the viewing direction satisfies
\begin{align}
\|\mathbf{v}(t)\|^2 = r_{02}^2(t) + r_{12}^2(t) + r_{22}^2(t) = 1.
\end{align}

Let $\mathbf{v}(t) = [v_x,\, v_y,\, v_z]^\top$. According to the spherical-coordinate formulation, these components are expressed as
\begin{equation}
\begin{aligned}
v_x &= \cos \phi(t)\, \sin \theta(t), \\
v_y &= \sin \phi(t), \\
v_z &= \cos \phi(t)\, \cos \theta(t).
\end{aligned}
\end{equation}
Without specifying a particular Euler-angle convention, the pitch and yaw angles are constrained within $\phi \in (-\tfrac{\pi}{2},\, \tfrac{\pi}{2}]$ and $\theta \in (-\pi,\, \pi]$. In practice, they are computed directly from the viewing-vector components as
\begin{align}
\scalebox{0.95}{$
\phi(t) = \arcsin(r_{12}(t)), \qquad
\theta(t) = \operatorname{atan2}(r_{02}(t),\, r_{22}(t)),
$}
\end{align}
where the two-argument arctangent ensures the correct quadrant for $\theta(t)$.
\subsection{Quantifying Acquisition Coverage and Discrete Morphological Adjustment}

To enable discrete and controllable sampling, we partition the viewing sphere in a longitude–latitude parameterization. The step sizes \(\Delta\theta\) and \(\Delta\phi\) fix the angular resolution: there are \(360^\circ/\Delta\theta\) longitudinal bins and \(180^\circ/\Delta\phi\) latitudinal bins, with both poles treated as boundary layers. This construction yields a uniformly spherical grid for indexing viewpoints.

The standard angular domain is $\theta'\in(-\pi,\pi]$ and $\phi'\in\left(-\tfrac{\pi}{2},\,\tfrac{\pi}{2}\right]$.To ensure continuity and prevent out-of-range values, each measured angle is normalized via wrapping:
\begin{align}
\theta' &= \operatorname{wrap}_{2\pi}(\theta)
= \theta - 2\pi \left\lfloor \frac{\theta + \pi}{2\pi} \right\rfloor, \\
\phi'   &= \operatorname{sat}_{\left(-\frac{\pi}{2},\,\frac{\pi}{2}\right]}(\phi).
\end{align}

\begin{figure}[t]
  \centering
  \includegraphics[width=0.75\linewidth]{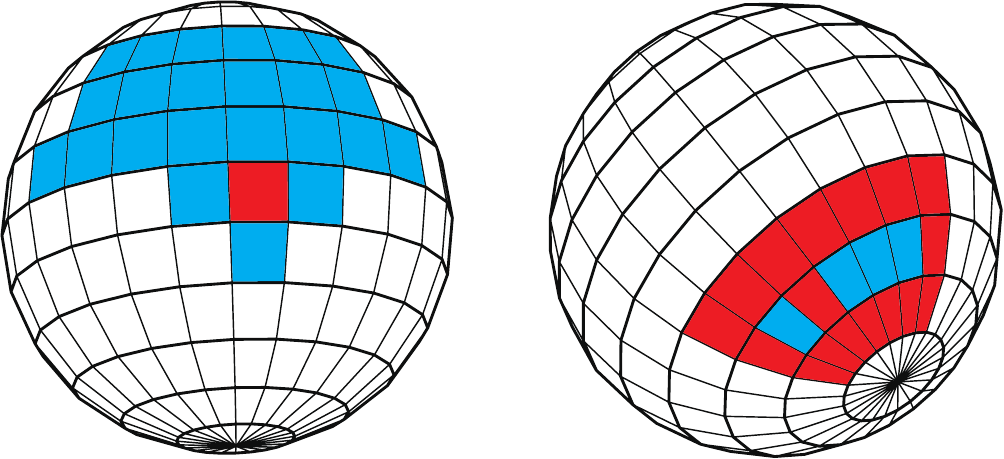}
  \caption{Morphological refinement of the discrete coverage map.
  \textbf{Blue}: original viewpoints; \textbf{Red}: added cells via adaptive dilation and hole filling. Improves local consistency without changing area-weighted coverage.}
  \label{fig:example}
\end{figure}

The normalized angles are then quantized into discrete grid indices according to the corresponding subdivisions:
\begin{align}
t &= \min\!\Bigl(N_{\theta}-1,\, \max\!\bigl(0, \bigl\lfloor \tfrac{\theta' + \pi}{2\pi}\, N_{\theta} \bigr\rfloor \bigr)\Bigr), \\
q &= \min\!\Bigl(N_{\phi}-1,\, \max\!\bigl(0, \bigl\lfloor \tfrac{\phi' + \frac{\pi}{2}}{\pi}\, N_{\phi} \bigr\rfloor \bigr)\Bigr).
\end{align}
For coverage bookkeeping, we denote the latitudinal index by $p$ (i.e., $p \equiv q$) when forming the binary coverage matrix.

Each grid point $(p,t)$ corresponds to an element in the binary coverage matrix $C \in \{0,1\}^{N_{\phi} \times N_{\theta}}$, which records acquisition states. 
Whenever an image is captured under a stable pose, the corresponding cell is updated as $C(p,t)\gets 1$. 
However, equal-angular discretization introduces a systematic bias, since cells near the poles represent smaller surface areas than those near the equator.

To correct this distortion, a spherical surface-area weighting is applied during coverage computation. 
The latitude sequence is defined as
\begin{align}
\phi\{p\} = -\frac{\pi}{2} + p\,\Delta\phi, \quad (p = 0, \ldots, N_{\phi}-1),
\end{align}
with longitudinal interval $\Delta\theta = \tfrac{2\pi}{N_{\theta}}$.
The differential area on the unit sphere is $dA = \cos\phi \, d\phi \, d\theta$. Accordingly,
\begin{equation}
\begin{aligned}
A_{p,t} &= \int_{\phi_p}^{\phi_{p+1}} \int_{\theta_t}^{\theta_{t+1}} 
\cos\phi \, d\theta \, d\phi \\
&= \Delta\theta \, \bigl(\sin \phi_{p+1} - \sin \phi_p\bigr).
\end{aligned}
\end{equation}

\begin{table*}[t]
\centering
\caption{Comparison of Free capture, RealityScan, and our method in terms of the number of captured images and reconstruction quality when scanning objects.~\cite{b15}~\cite{b16}~\cite{b17}
{\color{red}Red} and {\color{blue}blue} indicate the best and second-best performance, respectively. }
\label{tab:capture_quality_comparison}
\renewcommand{\arraystretch}{1}
\setlength{\tabcolsep}{3.2pt}
\footnotesize
\begin{tabular}{cccccccccccc}
\hline
Dataset & Methods & Times & No. of Images &
7K L1 & 7K PSNR & 7K SSIM & 7K LPIPS &
30K L1 & 30K PSNR & 30K SSIM & 30K LPIPS \\
\hline
& & {\color[HTML]{0D0D0D} 1} & {\color[HTML]{0D0D0D} 316} & {\color[HTML]{0D0D0D} 0.026} & {\color[HTML]{0D0D0D} 26.330} & {\color[HTML]{0D0D0D} 0.914} & {\color[HTML]{0D0D0D} 0.290} & {\color[HTML]{0D0D0D} 0.023} & {\color[HTML]{0D0D0D} 27.121} & {\color[HTML]{0D0D0D} 0.921} & {\color[HTML]{0D0D0D} 0.267} \\
& & {\color[HTML]{0D0D0D} 2} & {\color[HTML]{0D0D0D} 361} & 0.036 & {\color[HTML]{0D0D0D} 25.883} & {\color[HTML]{0D0D0D} 0.934} & {\color[HTML]{0D0D0D} 0.262} & {\color[HTML]{0D0D0D} 0.029} & {\color[HTML]{0D0D0D} 27.490} & {\color[HTML]{0D0D0D} 0.941} & {\color[HTML]{0D0D0D} 0.252} \\
& & {\color[HTML]{0D0D0D} 3} & {\color[HTML]{0D0D0D} 362} & {\color[HTML]{0D0D0D} 0.042} & {\color[HTML]{0D0D0D} 23.268} & {\color[HTML]{0D0D0D} 0.925} & {\color[HTML]{0D0D0D} 0.295} & {\color[HTML]{0D0D0D} 0.031} & {\color[HTML]{0D0D0D} 25.129} & {\color[HTML]{0D0D0D} 0.938} & {\color[HTML]{0D0D0D} 0.271} \\
& \multirow{-4}{*}{Free} & {\color[HTML]{0D0D0D} \textbf{Avg1-3}} & {\color[HTML]{0D0D0D} \textbf{346}} & {\color[HTML]{0D0D0D} \textbf{0.034}} & {\color[HTML]{0D0D0D} \textbf{25.160}} & {\color[HTML]{0D0D0D} \textbf{0.924}} & {\color[HTML]{0D0D0D} \textbf{0.282}} & {\color[HTML]{0D0D0D} \textbf{0.028}} & {\color[HTML]{0D0D0D} \textbf{26.580}} & {\color[HTML]{0D0D0D} \textbf{0.933}} & {\color[HTML]{0D0D0D} \textbf{0.263}} \\
& & {\color[HTML]{0D0D0D} 1} & {\color[HTML]{0D0D0D} 300} & {\color[HTML]{0D0D0D} 0.026} & {\color[HTML]{0D0D0D} 27.602} & {\color[HTML]{0D0D0D} 0.949} & {\color[HTML]{0D0D0D} 0.252} & {\color[HTML]{0D0D0D} 0.020} & {\color[HTML]{0D0D0D} 29.611} & {\color[HTML]{0D0D0D} 0.954} & {\color[HTML]{0D0D0D} 0.241} \\
& & {\color[HTML]{0D0D0D} 2} & {\color[HTML]{0D0D0D} 300} & {\color[HTML]{0D0D0D} 0.026} & {\color[HTML]{0D0D0D} 28.977} & {\color[HTML]{0D0D0D} 0.955} & {\color[HTML]{0D0D0D} 0.236} & {\color[HTML]{0D0D0D} 0.021} & {\color[HTML]{0D0D0D} 30.839} & {\color[HTML]{0D0D0D} 0.960} & {\color[HTML]{0D0D0D} 0.224} \\
& & {\color[HTML]{0D0D0D} 3} & {\color[HTML]{0D0D0D} 300} & {\color[HTML]{0D0D0D} 0.027} & {\color[HTML]{0D0D0D} 27.270} & {\color[HTML]{0D0D0D} 0.949} & {\color[HTML]{0D0D0D} 0.248} & {\color[HTML]{0D0D0D} 0.024} & {\color[HTML]{0D0D0D} 28.545} & {\color[HTML]{0D0D0D} 0.954} & {\color[HTML]{0D0D0D} 0.245} \\
& \multirow{-4}{*}{RealityScan} & {\color[HTML]{0D0D0D} \textbf{Avg1-3}} & {\color[HTML]{0D0D0D} \color[HTML]{00B0F0}\textbf{300}} & {\color[HTML]{00B0F0} \textbf{0.026}} & {\color[HTML]{00B0F0} \textbf{27.950}} & {\color[HTML]{FF0000} \textbf{0.951}} & {\color[HTML]{FF0000} \textbf{0.245}} & {\color[HTML]{00B0F0} \textbf{0.022}} & {\color[HTML]{00B0F0} \textbf{29.665}} & {\color[HTML]{FF0000} \textbf{0.956}} & {\color[HTML]{00B0F0} \textbf{0.237}} \\
& & {\color[HTML]{0D0D0D} 1} & {\color[HTML]{0D0D0D} 231} & {\color[HTML]{0D0D0D} 0.025} & {\color[HTML]{0D0D0D} 28.793} & {\color[HTML]{0D0D0D} 0.948} & {\color[HTML]{0D0D0D} 0.259} & {\color[HTML]{0D0D0D} 0.016} & {\color[HTML]{0D0D0D} 31.456} & {\color[HTML]{0D0D0D} 0.960} & {\color[HTML]{0D0D0D} 0.232} \\
& & {\color[HTML]{0D0D0D} 2} & {\color[HTML]{0D0D0D} 233} & {\color[HTML]{0D0D0D} 0.020} & {\color[HTML]{0D0D0D} 29.987} & {\color[HTML]{0D0D0D} 0.953} & {\color[HTML]{0D0D0D} 0.238} & {\color[HTML]{0D0D0D} 0.016} & {\color[HTML]{0D0D0D} 31.564} & {\color[HTML]{0D0D0D} 0.959} & {\color[HTML]{0D0D0D} 0.221} \\
& & {\color[HTML]{0D0D0D} 3} & {\color[HTML]{0D0D0D} 209} & {\color[HTML]{0D0D0D} 0.023} & {\color[HTML]{0D0D0D} 28.486} & {\color[HTML]{0D0D0D} 0.935} & {\color[HTML]{0D0D0D} 0.276} & {\color[HTML]{0D0D0D} 0.016} & {\color[HTML]{0D0D0D} 31.075} & {\color[HTML]{0D0D0D} 0.945} & {\color[HTML]{0D0D0D} 0.252} \\
\multirow{-12}{*}{Coinbank} & \multirow{-4}{*}{Ours} & {\color[HTML]{0D0D0D} \textbf{Avg1-3}} & {\color[HTML]{FF0000} \textbf{224}} & {\color[HTML]{FF0000} \textbf{0.023}} & {\color[HTML]{FF0000} \textbf{29.088}} & {\color[HTML]{00B0F0} \textbf{0.945}} & {\color[HTML]{00B0F0} \textbf{0.257}} & {\color[HTML]{FF0000} \textbf{0.016}} & {\color[HTML]{FF0000} \textbf{31.365}} & {\color[HTML]{00B0F0} \textbf{0.955}} & {\color[HTML]{FF0000} \textbf{0.235}} \\
\hline
& & 1 & 296 & 0.023 & 26.960 & 0.935 & 0.250 & 0.021 & 27.820 & 0.938 & 0.239 \\
& & 2 & 301 & 0.048 & 23.929 & 0.905 & 0.316 & 0.040 & 25.528 & 0.916 & 0.294 \\
& & 3 & 305 & 0.029 & 25.519 & 0.913 & 0.308 & 0.024 & 26.794 & 0.924 & 0.279 \\
& \multirow{-4}{*}{Free} & \textbf{Avg1-3} & \textbf{301} & {\color[HTML]{00B0F0} \textbf{0.034}} & \textbf{25.469} & \textbf{0.918} & \textbf{0.291} & {\color[HTML]{00B0F0} \textbf{0.028}} & \textbf{26.714} & \textbf{0.926} & \textbf{0.271} \\
& & 1 & 300 & 0.024 & 28.540 & 0.947 & 0.257 & 0.021 & 29.698 & 0.950 & 0.251 \\
& & 2 & 300 & 0.057 & 24.177 & 0.933 & 0.295 & 0.049 & 25.417 & 0.940 & 0.287 \\
& & 3 & 300 & 0.035 & 26.902 & 0.952 & 0.244 & 0.030 & 28.112 & 0.957 & 0.244 \\
& \multirow{-4}{*}{RealityScan} & \textbf{Avg1-3} & \color[HTML]{00B0F0}\textbf{300} & \textbf{0.039} & {\color[HTML]{00B0F0} \textbf{26.540}} & {\color[HTML]{00B0F0} \textbf{0.944}} & {\color[HTML]{00B0F0} \textbf{0.265}} & \textbf{0.034} & {\color[HTML]{00B0F0} \textbf{27.742}} & {\color[HTML]{00B0F0} \textbf{0.949}} & {\color[HTML]{00B0F0} \textbf{0.261}} \\
& & 1 & 237 & 0.025 & 27.590 & 0.924 & 0.297 & 0.025 & 27.590 & 0.924 & 0.297 \\
& & 2 & 248 & 0.009 & 35.639 & 0.972 & 0.213 & 0.004 & 45.123 & 0.979 & 0.164 \\
& & 3 & 235 & 0.030 & 27.023 & 0.941 & 0.266 & 0.020 & 30.130 & 0.952 & 0.241 \\
\multirow{-12}{*}{Terracotta Warrior model} & \multirow{-4}{*}{Ours} & \textbf{Avg1-3} & \color[HTML]{FF0000}\textbf{240} & {\color[HTML]{FF0000} \textbf{0.022}} & {\color[HTML]{FF0000} \textbf{30.084}} & {\color[HTML]{FF0000} \textbf{0.946}} & {\color[HTML]{FF0000} \textbf{0.259}} & {\color[HTML]{FF0000} \textbf{0.016}} & {\color[HTML]{FF0000} \textbf{34.281}} & {\color[HTML]{FF0000} \textbf{0.952}} & {\color[HTML]{FF0000} \textbf{0.234}} \\
\hline
& & 1 & 286 & 0.019 & 28.571 & 0.932 & 0.223 & 0.015 & 30.358 & 0.946 & 0.192 \\
& & 2 & 312 & 0.046 & 22.321 & 0.871 & 0.353 & 0.033 & 24.463 & 0.896 & 0.305 \\
& & 3 & 303 & 0.034 & 24.556 & 0.894 & 0.292 & 0.031 & 25.272 & 0.901 & 0.269 \\
& \multirow{-4}{*}{Free} & \textbf{Avg1-3} & \textbf{300} & \textbf{0.033} & \textbf{25.149} & \textbf{0.899} & {\color[HTML]{00B0F0} \textbf{0.289}} & \textbf{0.026} & \textbf{26.698} & {\color[HTML]{00B0F0} \textbf{0.914}} & {\color[HTML]{00B0F0} \textbf{0.255}} \\
& & 1 & 300 & 0.028 & 26.465 & 0.911 & 0.265 & 0.026 & 27.090 & 0.914 & 0.246 \\
& & 2 & 300 & 0.028 & 25.830 & 0.901 & 0.301 & 0.023 & 27.209 & 0.915 & 0.264 \\
& & 3 & 300 & 0.033 & 24.837 & 0.899 & 0.307 & 0.027 & 26.535 & 0.915 & 0.277 \\
& \multirow{-4}{*}{RealityScan} & \textbf{Avg1-3} & \color[HTML]{00B0F0}\textbf{300} & {\color[HTML]{00B0F0} \textbf{0.030}} & {\color[HTML]{00B0F0} \textbf{25.711}} & {\color[HTML]{00B0F0} \textbf{0.904}} & \textbf{0.291} & {\color[HTML]{00B0F0} \textbf{0.025}} & {\color[HTML]{00B0F0} \textbf{26.945}} & \textbf{0.915} & \textbf{0.262} \\
& & 1 & 288 & 0.028 & 25.478 & 0.911 & 0.284 & 0.025 & 26.329 & 0.918 & 0.260 \\
& & 2 & 237 & 0.006 & 40.546 & 0.969 & 0.169 & 0.005 & 43.790 & 0.975 & 0.142 \\
& & 3 & 250 & 0.038 & 23.213 & 0.883 & 0.402 & 0.028 & 25.465 & 0.904 & 0.354 \\
\multirow{-12}{*}{Miniclawmachine} & \multirow{-4}{*}{Ours} & \textbf{Avg1-3} & \color[HTML]{FF0000}\textbf{258} & {\color[HTML]{FF0000} \textbf{0.024}} & {\color[HTML]{FF0000} \textbf{29.746}} & {\color[HTML]{FF0000} \textbf{0.921}} & {\color[HTML]{FF0000} \textbf{0.285}} & {\color[HTML]{FF0000} \textbf{0.019}} & {\color[HTML]{FF0000} \textbf{31.861}} & {\color[HTML]{FF0000} \textbf{0.932}} & {\color[HTML]{FF0000} \textbf{0.252}} \\
\hline
\end{tabular}
\end{table*}

Based on this area element, the area-weighted coverage rate is defined as
\begin{align}
C_{\mathrm{area}} =
\frac{
\sum_{p=0}^{N_{\phi}-2} \sum_{t=0}^{N_{\theta}-1} C(p,t)\, A_{p,t}
}{
\sum_{p=0}^{N_{\phi}-2} \sum_{t=0}^{N_{\theta}-1} A_{p,t}
} \times 100\%.
\end{align}
This formulation is consistent with the spherical surface measure, avoiding overestimation of grid density near the poles. The parameters are chosen through multiple experiments to balance capture usability and reconstruction quality.

To address numerical instability and discretization near the poles, we apply two morphological refinements to the coverage matrix $C$. 
Adaptive pole dilation expands sparse polar regions within a predefined zone while preserving longitudinal continuity. 
Hole filling removes discretization voids by activating cells whose four axial neighbors are active.

\section{Results and Analysis}

\begin{figure}[t]
  \centering
  \includegraphics[width=0.8\linewidth]{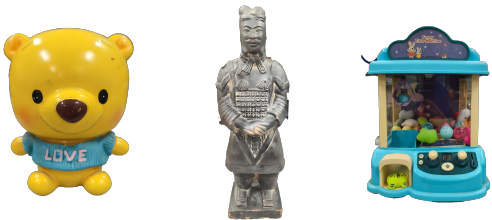}
  \caption{Evaluation Objects: Coinbank, Terracotta Warrior model, Miniclawmachine (from left to right)}
  \label{fig:coinbank}
\end{figure}

Experiments utilize a Redmi K70 Pro for capture and an NVIDIA RTX 5090D GPU for off-device 3DGS reconstruction. The test set comprises three tabletop objects in Fig.~\ref{fig:coinbank}.

\subsection{Performance Evaluation}

We compare our method with free capture without guidance and RealityScan~\cite{b14}. RealityScan is an automated photogrammetric acquisition and reconstruction system that triggers image capture by evaluating the geometric novelty of the current viewpoint relative to previously captured images. Views with similar camera poses and minimal visual change are discarded, resulting in an implicitly guided acquisition process that encourages view diversity without explicit user feedback. The number of images captured using RealityScan is determined by its default system settings. Furthermore, free capture here is implemented using our acquisition framework with the visual guidance module disabled. No explicit guidance is provided to the user, while camera poses and acquisition trajectories are still recorded for evaluation purposes. As shown in Table~\ref{tab:capture_quality_comparison}, where 7K and 30K denote the training iterations steps of 3DGS, although our method captures fewer images than both free capture and RealityScan, the acquired views are more effective and with lower redundancy, leading to consistently advanced reconstruction quality across all evaluation metrics.

\subsection{Coverage Evaluation}

\begin{table}
\centering
\footnotesize
\caption{Band-wise comparison of image counts (\textbf{Img.}) and coverage (\textbf{Cov.}) between free capture and our method for Coinbank. \textbf{Avg.} denotes the average of free capture. Left to right: front ($-45^\circ$ to $45^\circ$), side, back, and opposite side.}

\label{tab:bandwise_all}
\setlength{\tabcolsep}{2pt}
\renewcommand{\arraystretch}{0.95}
\resizebox{\columnwidth}{!}{%
\begin{tabular}{lcccccccccc}
\hline
                       & \multicolumn{2}{c}{$-45^\circ$--$45^\circ$} 
                       & \multicolumn{2}{c}{$45^\circ$--$135^\circ$} 
                       & \multicolumn{2}{c}{$135^\circ$--$-135^\circ$} 
                       & \multicolumn{2}{c}{$-135^\circ$--$45^\circ$} 
                       & \multicolumn{2}{c}{PSNR (dB)} \\
\multirow{-2}{*}{Test} & Img. & Cov. 
                       & Img. & Cov.
                       & Img. & Cov.
                       & Img. & Cov.
                       & 7K & 30K \\ \hline
1  & 120 & 73\% & 25 & 44\% & 51 & 40\% & 42 & 52\% & 25.920 & 28.774 \\
2  & 90  & 69\% & 50 & 57\% & 79 & 52\% & 70 & 73\% & 25.234 & 26.950 \\
3  & 109 & 48\% & 63 & 61\% & 46 & 57\% & 81 & 48\% & 26.864 & 28.888 \\
4  & 63  & 65\% & 71 & 61\% & 50 & 32\% & 74 & 52\% & 25.665 & 26.810 \\
5  & 92  & 61\% & 88 & 60\% & 93 & 52\% & 57 & 61\% & 25.611 & 27.024 \\
6  & 159 & 67\% & 85 & 46\% & 56 & 33\% & 52 & 47\% & 26.786 & 27.341 \\
7  & 140 & 67\% & 83 & 40\% & 29 & 33\% & 65 & 43\% & 26.659 & 28.846 \\
8  & 170 & 63\% & 67 & 47\% & 32 & 33\% & 47 & 50\% & 26.330 & 27.121 \\
9  & 108 & 53\% & 35 & 27\% & 69 & 37\% & 70 & 50\% & 28.106 & 29.771 \\
10 & 178 & 73\% & 91 & 47\% & 50 & 43\% & 42 & 43\% & 25.883 & 27.490 \\ \hline
$\mathbf{Avg}$ 
   & \textbf{123} & \textit{\textbf{64\%}} 
   & \textbf{66}  & \textit{\textbf{49\%}} 
   & \textbf{56}  & \textit{\textbf{41\%}} 
   & \textbf{60}  & \textit{\textbf{52\%}} 
   & \textbf{26.306} & \textbf{27.902} \\

\textbf{Ours} 
   & \textbf{82} & \textit{\textbf{100\%}} 
   & \textbf{53} & \textit{\textbf{100\%}} 
   & \textbf{63} & \textit{\textbf{100\%}} 
   & \textbf{60} & \textit{\textbf{100\%}} 
   & \textbf{26.779} & \textbf{30.276} \\ \hline
\end{tabular}%
}
\end{table}

\begin{figure}[t]
  \centering
  \includegraphics[width=0.65\columnwidth]{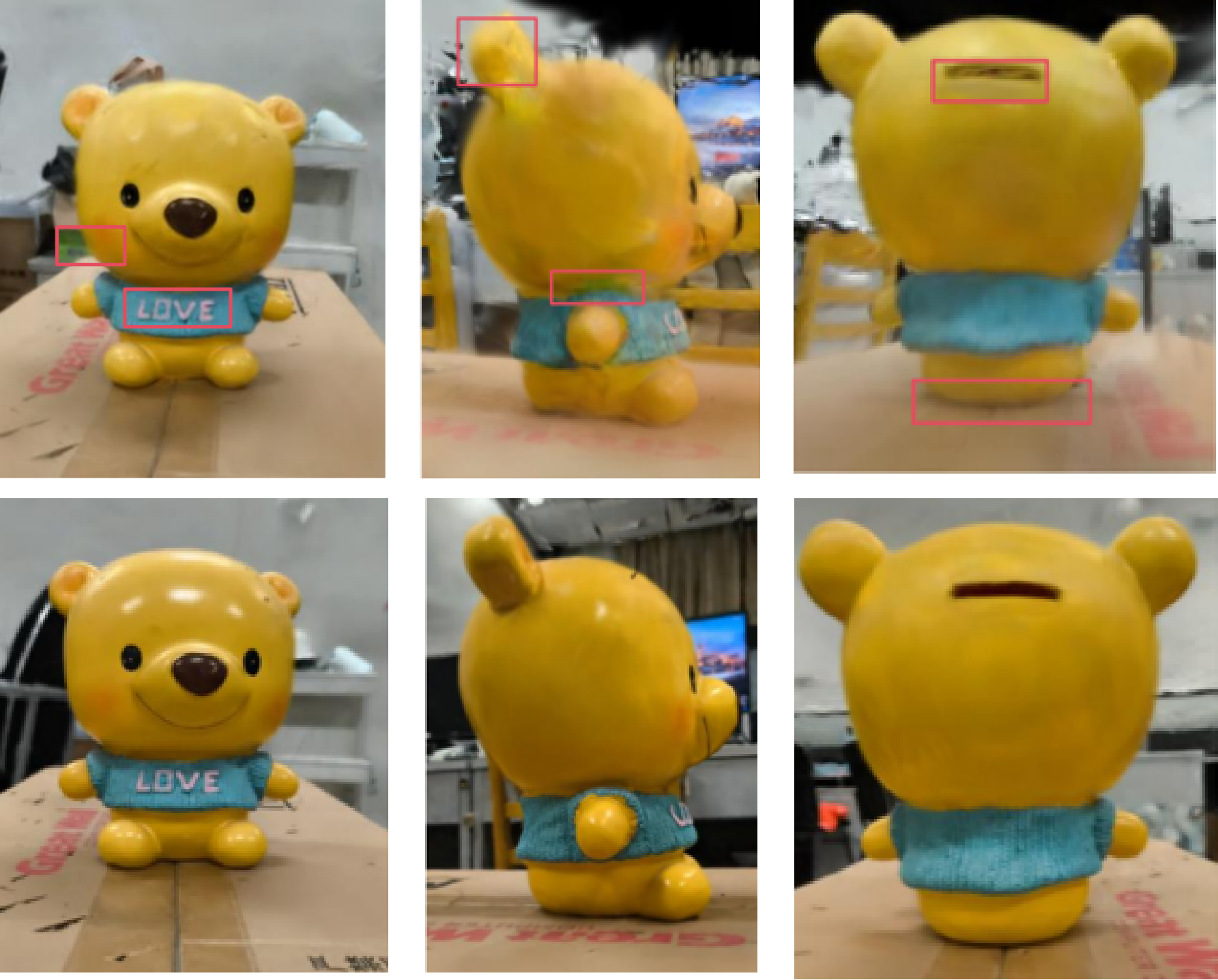}
  \caption{Side-by-side reconstructions at identical viewpoints for free(top row) vs. guided acquisitions for the  Coinbank.
}
  \label{fig:efficiency_psnr}
\end{figure}

A user study with ten participants on Coinbank is conducted to assess the impact of acquisition guidance on viewpoint coverage and data efficiency, comparing grid-guided capture against free capture. It is noted that RealityScan does not provide the related coverage information, so it is not included in Table~\ref{tab:bandwise_all}.
In the free capture setting, the proposed acquisition framework is used with the guidance module disabled, so the images are captured based on the judgement of the subjective user. The viewpoint-related information is recorded only for evaluation purposes. The quantitative results are reported in Table~\ref{tab:bandwise_all}. It is seen that free capture exhibits non-uniform coverage across the four horizontal angle $\theta$ partitions, each spanning $90^\circ$. Coverage is highest in the frontal region and decreases in subsequent regions, which can lead to reconstruction artifacts as shown in Fig.~\ref{fig:efficiency_psnr}, then increasing again as viewpoints return toward the frontal area.

\begin{figure}[t]
  \centering
  \includegraphics[width=0.7\linewidth]{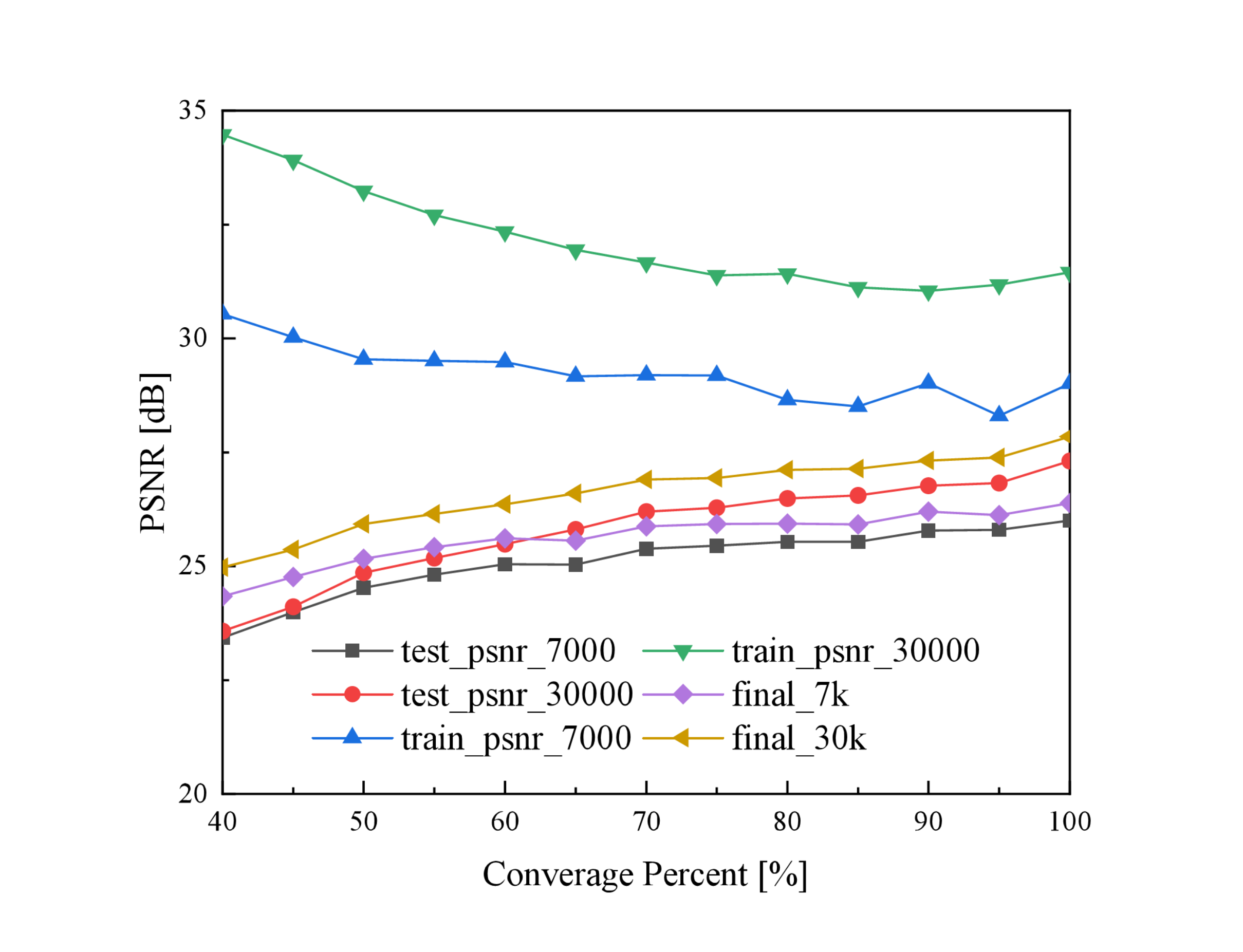}
  \caption{Ablation Study: PSNR versus coverage on Coinbank.}
  \label{fig:coverage_psnr_ablation}
\end{figure}

To analyze how viewpoint coverage affects reconstruction quality, we  progressively reduce training data for 3DGS. We project training viewpoints onto a spherical grid to compute area-weighted coverage and randomly remove training images while keeping the evaluation set fixed. As shown in Fig.~\ref{fig:coverage_psnr_ablation}, reconstruction quality degrades as coverage decreases.

\subsection{Ablation Study}

\begin{table}[t]
\centering
\footnotesize
\caption{Ablation study on camera pose refinement. PSNR comparison between original COLMAP poses and camera poses updated using device-computed orientations aligned with the object center on (A) Coinbank, (B) Miniclawmachine (C) Terracotta Warrior. \textcolor{red}{Red} means better performance.}
\label{tab:psnr_coinbank}
\setlength{\tabcolsep}{4pt}
\renewcommand{\arraystretch}{0.9}
\begin{adjustbox}{max width=\columnwidth}
\begin{tabular}{lcccc}
\hline
\multirow{2}{*}{Scene} & \multicolumn{2}{c}{Origin PSNR (dB)} & \multicolumn{2}{c}{Refined PSNR (dB)} \\
 & 7k & 30k & 7k & 30k \\ \hline
\multirow{3}{*}{Obj.A}
 & 27.121 & 30.007 & \textcolor{red}{27.330} & \textcolor{red}{30.312} \\
 & 27.094 & 29.941 & \textcolor{red}{27.293} & \textcolor{red}{30.161} \\
 & 26.967 & 30.134 & \textcolor{red}{27.330} & \textcolor{red}{30.331} \\\hline 
\multirow{3}{*}{Obj.B}
 & 24.952 & 27.573 & \textcolor{red}{25.116} & \textcolor{red}{27.634} \\
 & 24.818 & 27.574 & \textcolor{red}{25.002} & \textcolor{red}{27.678} \\
 & 25.027 & \textcolor{red}{27.744} & \textcolor{red}{25.120} & 27.593 \\\hline 
\multirow{3}{*}{Obj.C}
 & 26.703 & 28.680 & \textcolor{red}{26.813} & \textcolor{red}{28.685} \\
 & 26.726 & 28.505 & \textcolor{red}{26.735} & \textcolor{red}{28.638} \\
 & 26.716 & 28.638 & \textcolor{red}{26.799} & \textcolor{red}{28.813} \\ \hline
\end{tabular}
\end{adjustbox}
\end{table}

Since the proposed method computes camera orientations relative to the object center during acquisition provided with the sensor data, we evaluate the effectiveness for using the real pose information for 3DGS rendering. Specifically, we compare the results when the camera centers and all other pose information is estimated by COLMAP~\cite{b18} (denoted as COLMAP), to those using the proposed computation of orientations aligned to the object center using the real data from the device (denoted as Proposed). The related experiments are carried out  under identical training settings. As shown in Table~\ref{tab:psnr_coinbank}, real data from sensors are useful for improving 3DGS reconstruction quality as proposed in our work.

\section{Conclusion}
We present an object-centered, real-time mobile capture method that integrates uniform viewpoint indexing with IMU-guided online coverage estimation. The proposed approach enables more uniform and complete multi-view acquisition under handheld conditions, consistently improving reconstruction quality while requiring fewer input images.
We may incorporate additional distance constraints in future work.

\newpage
\nocite{*}
\bibliographystyle{IEEEbib}
\bibliography{refs}

@article{b1,
  title={3d gaussian splatting for real-time radiance field rendering.},
  author={Kerbl, Bernhard and Kopanas, Georgios and Leimk{\"u}hler, Thomas and Drettakis, George and others},
  journal={ACM Trans. Graph.},
  volume={42},
  number={4},
  pages={1--14},
  year={2023}
}

@article{b2,
  title={Neural radiance field-based visual rendering: A comprehensive review},
  author={Yao, Mingyuan and Huo, Yukang and Ran, Yang and Tian, Qingbin and Wang, Ruifeng and Wang, Haihua},
  journal={arXiv preprint arXiv:2404.00714},
  year={2024}
}

@article{b3,
  title={Instant neural graphics primitives with a multiresolution hash encoding},
  author={M{\"u}ller, Thomas and Evans, Alex and Schied, Christoph and Keller, Alexander},
  journal={ACM transactions on graphics (TOG)},
  volume={41},
  number={4},
  pages={1--15},
  year={2022},
  publisher={ACM New York, NY, USA}
}

@article{b4,
  title={Nerf: Representing scenes as neural radiance fields for view synthesis},
  author={Mildenhall, Ben and Srinivasan, Pratul P and Tancik, Matthew and Barron, Jonathan T and Ramamoorthi, Ravi and Ng, Ren},
  journal={Communications of the ACM},
  volume={65},
  number={1},
  pages={99--106},
  year={2021},
  publisher={ACM New York, NY, USA}
}

@inproceedings{b5,
  title={Regnerf: Regularizing neural radiance fields for view synthesis from sparse inputs},
  author={Niemeyer, Michael and Barron, Jonathan T and Mildenhall, Ben and Sajjadi, Mehdi SM and Geiger, Andreas and Radwan, Noha},
  booktitle={Proceedings of the IEEE/CVF conference on computer vision and pattern recognition},
  pages={5480--5490},
  year={2022}
}

@inproceedings{b6,
  title={4d gaussian splatting for real-time dynamic scene rendering},
  author={Wu, Guanjun and Yi, Taoran and Fang, Jiemin and Xie, Lingxi and Zhang, Xiaopeng and Wei, Wei and Liu, Wenyu and Tian, Qi and Wang, Xinggang},
  booktitle={Proceedings of the IEEE/CVF conference on computer vision and pattern recognition},
  pages={20310--20320},
  year={2024}
}

@article{b7,
  title={Vins-mono: A robust and versatile monocular visual-inertial state estimator},
  author={Qin, Tong and Li, Peiliang and Shen, Shaojie},
  journal={IEEE transactions on robotics},
  volume={34},
  number={4},
  pages={1004--1020},
  year={2018},
  publisher={IEEE}
}

@inproceedings{b8,
  title={Structure-from-motion revisited},
  author={Schonberger, Johannes L and Frahm, Jan-Michael},
  booktitle={Proceedings of the IEEE conference on computer vision and pattern recognition},
  pages={4104--4113},
  year={2016}
}

@inproceedings{b9,
  title={A multi-view stereo benchmark with high-resolution images and multi-camera videos},
  author={Schops, Thomas and Schonberger, Johannes L and Galliani, Silvano and Sattler, Torsten and Schindler, Konrad and Pollefeys, Marc and Geiger, Andreas},
  booktitle={Proceedings of the IEEE conference on computer vision and pattern recognition},
  pages={3260--3269},
  year={2017}
}

@inproceedings{b10,
  title={A multi-state constraint Kalman filter for vision-aided inertial navigation},
  author={Mourikis, Anastasios I and Roumeliotis, Stergios I},
  booktitle={Proceedings 2007 IEEE international conference on robotics and automation},
  pages={3565--3572},
  year={2007},
  organization={IEEE}
}

@inproceedings{b11,
  title={Unified temporal and spatial calibration for multi-sensor systems},
  author={Furgale, Paul and Rehder, Joern and Siegwart, Roland},
  booktitle={2013 IEEE/RSJ International Conference on Intelligent Robots and Systems},
  pages={1280--1286},
  year={2013},
  organization={IEEE}
}

@book{b12,
  title={Smoothing, forecasting and prediction of discrete time series},
  author={Brown, Robert Goodell},
  year={2004},
  publisher={Courier Corporation}
}

@article{b13,
  title={Nonlinear complementary filters on the special orthogonal group},
  author={Mahony, Robert and Hamel, Tarek and Pflimlin, Jean-Michel},
  journal={IEEE Transactions on automatic control},
  volume={53},
  number={5},
  pages={1203--1218},
  year={2008},
  publisher={IEEE}
}

@misc{b14,
  author = {{Epic Games}},
  title  = {RealityScan},
  year   = {2024},
  note   = {\url{https://www.realityscan.com/}}
}

@article{b15,
  title={Image quality assessment: from error visibility to structural similarity},
  author={Wang, Zhou and Bovik, Alan C and Sheikh, Hamid R and Simoncelli, Eero P},
  journal={IEEE transactions on image processing},
  volume={13},
  number={4},
  pages={600--612},
  year={2004},
  publisher={IEEE}
}

@article{b16,
  title={Advantages of the mean absolute error (MAE) over the root mean square error (RMSE) in assessing average model performance},
  author={Willmott, Cort J and Matsuura, Kenji},
  journal={Climate research},
  volume={30},
  number={1},
  pages={79--82},
  year={2005}
}

@inproceedings{b17,
  title={The unreasonable effectiveness of deep features as a perceptual metric},
  author={Zhang, Richard and Isola, Phillip and Efros, Alexei A and Shechtman, Eli and Wang, Oliver},
  booktitle={Proceedings of the IEEE conference on computer vision and pattern recognition},
  pages={586--595},
  year={2018}
}

@inproceedings{b18,
  title={Pixelwise view selection for unstructured multi-view stereo},
  author={Sch{\"o}nberger, Johannes L and Zheng, Enliang and Frahm, Jan-Michael and Pollefeys, Marc},
  booktitle={European conference on computer vision},
  pages={501--518},
  year={2016},
  organization={Springer}
}

\end{document}